%% file: acl2021.tex
\documentclass[11pt,a4paper]{article}
\usepackage[hyperref]{acl2021}
\usepackage{times}
\usepackage{latexsym}
\usepackage{pifont}
\usepackage{microtype}
\usepackage{amssymb}

\aclfinalcopy 
\def\aclpaperid{1740} 

\usepackage{microtype}
\usepackage{booktabs}
\usepackage{url}
\usepackage{eurosym}
\usepackage{todonotes}
\usepackage{multirow}
\usepackage{subcaption}
\usepackage{amssymb}
\usepackage{amsmath}
\usepackage{enumitem}
\usepackage{array}
\usepackage{longtable}
\usepackage{makecell}

\newcommand{\figref}[1]{Figure~\ref{fig:#1}}

\newcommand{\tabref}[1]{Table~\ref{tab:#1}}
\newcommand{\finqa}{\textsc{TAT-QA}}
\newcommand{\tagop}{\textsc{TagOp}}
\newcommand{\tapas}{\text{TaPas}}

\newcommand{\PreserveBackslash}[1]{\let\temp=\\#1\let\\=\temp}
\newcolumntype{C}[1]{>{\PreserveBackslash\centering}p{#1}}
\newcolumntype{R}[1]{>{\PreserveBackslash\raggedleft}p{#1}}
\newcolumntype{L}[1]{>{\PreserveBackslash\raggedright}p{#1}}

\newcommand\fone{F\textsubscript{1}}
\usepackage{microtype}

\title{\finqa: A Question Answering Benchmark on a Hybrid of Tabular and Textual Content in Finance}

\author{Fengbin Zhu\textsuperscript{1,2},~Wenqiang Lei\textsuperscript{1}$\thanks{$^{*}$Corresponding author}$,~Youcheng Huang \textsuperscript{3},~Chao Wang\textsuperscript{2}, Shuo Zhang\textsuperscript{4},\\\textbf{Jiancheng Lv}\textsuperscript{3},~\textbf{Fuli Feng}\textsuperscript{1},~\textbf{Tat-Seng Chua}\textsuperscript{1}
\\
\textsuperscript{1}National University of Singapore,~\textsuperscript{2}6Estates Pte Ltd,~\textsuperscript{3}Sichuan University,~\textsuperscript{4}Bloomberg\\

\tt{\{zhfengbin, wenqianglei\}@gmail.com},~\tt {wangchao@6estates.com}}

\date{}

\begin{document}
\maketitle
\begin{abstract}

Hybrid data combining both tabular and textual content (e.g., financial reports) are quite pervasive in the real world. 
However, Question Answering (QA) over such hybrid data is largely neglected in existing research.
In this work, we extract samples from real financial reports to build a new large-scale QA dataset containing both \textbf{T}abular \textbf{A}nd \textbf{T}extual data, named \finqa, where numerical reasoning is usually required to infer the answer, such as addition, subtraction, multiplication, division, counting, comparison/sorting, and their compositions.
We further propose a novel QA model termed \tagop, which is capable of reasoning over both tables and text.
It adopts sequence tagging to extract relevant cells from the table along with relevant spans from the text to infer their semantics, and then applies symbolic reasoning over them with a set of aggregation operators to arrive at the final answer. 
\tagop~achieves $58.0$\% in \fone{}, which is an $11.1$\% absolute increase over the previous best baseline model, according to our experiments on \finqa. 
But this result still lags far behind the performance of human expert, i.e. $90.8$\% in \fone{}.
It demonstrates that our \finqa~is very challenging and can serve as a benchmark for training and testing powerful QA models that address hybrid data.
Our dataset is publicly
available for non-commercial use at~\url{https://nextplusplus.github.io/TAT-QA/}.

\end{abstract}

\section{Introduction}
\label{sec:intro}
\input{01_intro}

\section{Dataset Construction and Analysis}
\label{sec:construction}
\input{02_data_col}

\section{\tagop~Model}
\label{sec:model}
\input{03_model}

\section{Experiments and Results}
\label{sec:experiments}
\input{04_experiment}

\section{Related Work}
\label{sec:related}
\input{05_related_work}

\section{Conclusion}
\label{sec:conclusion}
\input{06_conclusion}

\section*{Acknowledgments}
The authors gratefully acknowledge Zhuyun Dai for giving valuable suggestions on this study, Xinnan Zhang for developing the data annotation tool, and Tong Ye and Ming Wei Chan for their work on checking the annotation quality.
Our thanks also go to all the anonymous reviewers for their positive feedback. 
This research is supported by the NExT Research Centre, Singapore.

\bibliography{acl2021}
\bibliographystyle{acl_natbib}

\appendix
\section{Appendix}
\label{sec:appendix}
\input{07_appendix}

\end{document}

%% file: 01_intro.tex
Existing QA systems largely focus on only unstructured 
text~\cite{Hermann2015Teaching,Rajpurkar2016SQuAD,Dua2019DROP,yang2018hotpotqa,lei2020molweni,nie2020large}, structured knowledge base (KB)~\citep{berant2013semantic,yih2015semantic,talmor2018web}, or semi-structured tables~\citep{Pasupat2015Compositional,zhongSeq2SQL2017,yu2018spider,Zhang2019ADC,Zhang2020SET}.
Though receiving growing interests~\citep{das2017question,Sun2019Pullnet,chen2020hybridqa,chen2021open}, works on hybrid data comprising of unstructured text and structured or semi-structured KB/tables are rare.
Recently, \citet{chen2020hybridqa} attempt to simulate a type of hybrid data through manually linking table cells to Wiki pages via hyperlinks. 
However, such connection between table and text is relatively loose.

\begin{figure*}[htb]
    \begin{center}
    \includegraphics[scale=0.538]{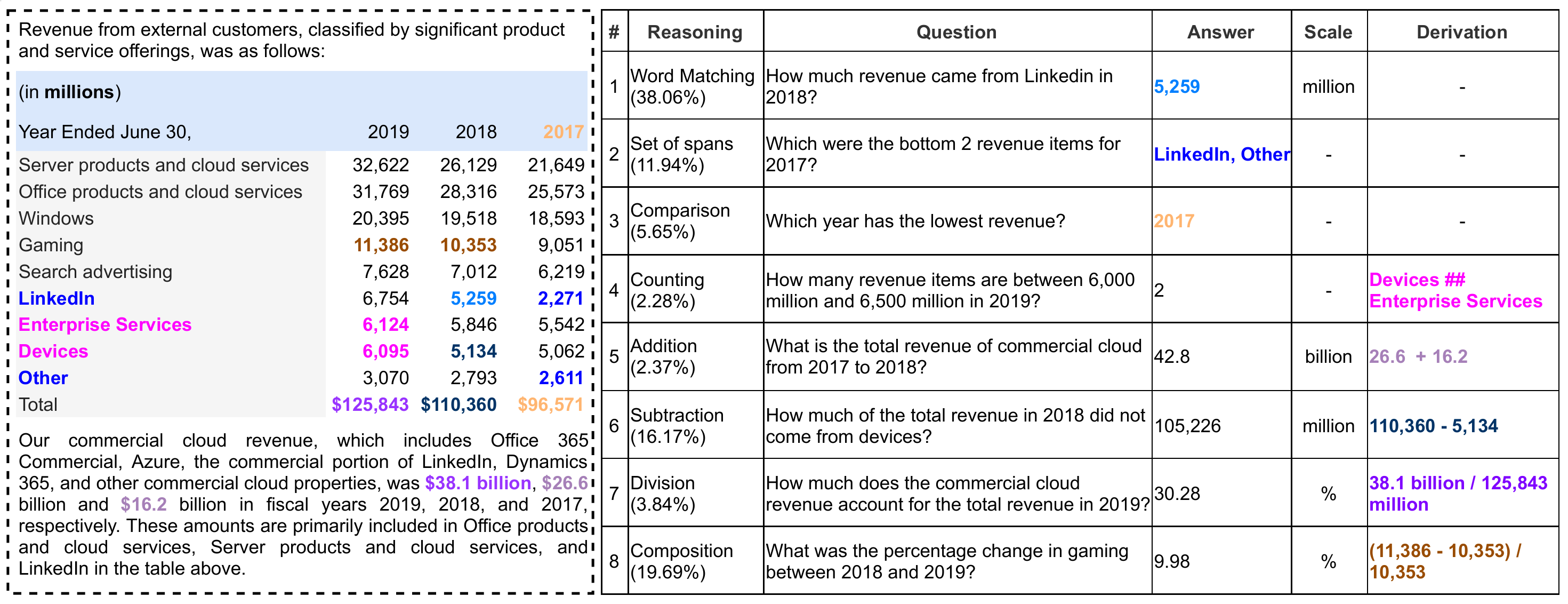}
    \end{center}
    \vspace{-1em}
    \caption{\label{fig:sample} 
    An example of \finqa. 
    The left dashed line box shows a \textit{hybrid context}.
    The rows with blue background are \textit{row header} while the column with grey is \textit{column header}.
    The right solid line box shows corresponding question, answer with its scale,  and derivation to arrive at the answer.
   }
\end{figure*}

In the real world, a more common hybrid data form is, the table (that usually contains numbers) is more comprehensively linked to text, e.g., semantically related or complementary.
Such hybrid data are very pervasive in various scenarios like scientific research papers, medical reports, financial reports, etc.
The left box of \figref{sample} shows a real example from some financial report, where there is a table containing row/column header and numbers inside, and also some paragraphs describing it.
We call the hybrid data like this example \textit{hybrid context}  in QA problems, as it contains both tabular and textual content, and call the paragraphs \textit{associated paragraphs} to the table.
To comprehend and answer a question from such hybrid context relies on the close relation between table and paragraphs, and usually requires numerical reasoning.
For example, one needs to identify ``revenue from the external customers'' in the describing text so as to understand the content of the table. 
As for \emph{``How much does the commercial cloud revenue account for the total revenue in 2019?''}, one needs to get the total revenue in 2019, i.e. ``$125,843$ million'' from the table and commercial cloud revenue, i.e. ``$38.1$ billion'', from the text to infer the answer.

To stimulate progress of QA research over such hybrid data, we propose a new dataset, named \finqa~(\textbf{T}abular \textbf{A}nd \textbf{T}extual dataset for \textbf{Q}uestion \textbf{A}nswering). 
The hybrid contexts in \finqa~are extracted from real-world financial reports, 
each composed of a table with row/col header and numbers, as well as at least two paragraphs that describe, analyse or complement the content of this table. 
Given hybrid contexts, we invite annotators with financial knowledge to generate questions that are useful in real-world financial analyses and provide answers accordingly.
It is worth mentioning that a large portion of questions in \finqa~demand numerical reasoning, for which derivation of the answer is also labeled to facilitate developing explainable models.
In total, \finqa~contains $16,552$ questions associated with $2,757$ hybrid contexts from $182$ reports.

We further propose a novel \tagop~model based on \finqa.
Taking as input the given question, table and associated paragraphs, \tagop~applies sequence tagging to extract relevant cells from the table and relevant spans from text as the evidences.
Then it applies symbolic reasoning over them with a set of aggregation operators to arrive at the final answer.
Predicting the magnitude of a number is an important aspect when tackling hybrid data in \finqa, including thousand, million, billion, etc. that are often omitted or shown only in headers or associated paragraphs of the table for brevity.
We term such magnitude of a number as its \textit{scale}.
Take Question 6 in \figref{sample} as an example:
\emph{``How much of the total revenue in 2018 did not come from devices?''}
The numerical value in the answer is obtained by subtraction: ``$110,360$ - $5,134$'', while the scale ``million'' is identified from the first-row header of the table.
In \tagop, we incorporate a multi-class classifier for scale prediction.

We test three types of QA models on \finqa, specially addressing tabular, textual, and hybrid data. 
Our \tagop~achieves $58.0$\% in terms of \fone{}, which is a $11.1$\% absolute increase over the best baseline model, according to our experiments on \finqa.
It is worth noting that the results still lag far behind performance of human experts, i.e. $90.8$\% in \fone{}.
We can see that to tackle the QA task over the hybrid data as in \finqa~is challenging and more effort is demanded.
We expect our \finqa~dataset and \tagop~model to serve as a benchmark and baseline respectively to contribute to the development of QA models for hybrid data, especially those requiring numerical reasoning.

%% file: 02_data_col.tex
We here explain how we construct \finqa~ and analyze its statistics to better reveal its proprieties.

\subsection{Data Collection and Preprocessing}

In \finqa~there are two forms of data: tables and their relevant text, which are extracted from real-world financial reports. 

In particular, we first download about $500$ financial reports released in the past two years from an online website\footnote{https://www.annualreports.com/}.
We adopt the table detection model in~\citep{li2019tablebank} to detect tables in these reports, and apply Apache PDFBox\footnote{https://pdfbox.apache.org/} library to extract the table contents to be processed with our annotation tool.
We only keep those tables with $3$ $\sim$ $30$ rows and $3$ $\sim$ $6$ columns.
Finally, about $20,000$ candidate tables are retained, which have no standard schema and lots of numbers inside.
The corresponding reports with selected tables are also kept.
Note that these candidate tables may still contain errors, such as containing too few or many rows/cols, mis-detected numbers, which will be manually picked out and deleted or fixed during the annotation process.  

\subsection{Dataset Annotation}
The annotation is done with our self-developed tool.
All the annotators are with financial background knowledge.

\noindent \textbf{Adding Relevant Paragraphs to Tables}
We build valid hybrid contexts based on the original reports kept in the previous step.
A valid hybrid context in \finqa~consists of a table and at least two associated paragraphs surrounding it, as shown in the left  box in \figref{sample}.
To associate enough relevant paragraphs to a candidate table, the annotators first check whether there are $\ge2$ paragraphs around this table, and then check whether they are relevant, meaning the paragraphs should be describing, analysing or complementing the content in the table. 
If yes, then all the surrounding paragraphs will be associated to this table.
Otherwise, the table will be skipped (discarded).\footnote{About two thirds of candidate tables were discarded.}

\noindent \textbf{Question-Answer Pair Creation} 
Based on the valid hybrid contexts, the annotators are then asked to create question-answer pairs, where the questions need to be useful in real-world financial analyses. 
In addition, we encourage them to create questions that can be answered by people without much finance knowledge and use common words instead of the same words appeared in the hybrid context~\citep{Rajpurkar2016SQuAD}.
Given one hybrid context, at least $6$ questions are generated, including \textit{extracted} and \textit{calculated} questions.
For \textit{extracted} questions, the answers can be a single span or multiple spans from either the table or the associated paragraphs.
For \textit{calculated} questions, numerical reasoning is required to produce the answers, 
including addition, subtraction, multiplication, division, counting, comparison/sorting and their compositions.
Furthermore, we particularly ask the annotators to annotate the right scale for the numerical answer when necessary.

\noindent \textbf{Answer Type and Derivation Annotation}
The answers in \finqa~have three types: a single span or multiple spans extracted from the table or text, as well as a generated answer (usually obtained through numerical reasoning).
The annotators will also need to label its type after they generate an answer.
For generated answers, the corresponding derivations are provided to facilitate the development of explainable QA models, including two types: 
1) an arithmetic expression,  like  \emph{($11,386$ - $10,353$)/$10,353$)} for Question 8 in \figref{sample}, which can be executed to arrive at the final answer;
and 2) a set of items separated with \emph{``\#\#''}, like \emph{``device \#\# enterprise services''} for Question 4 in \figref{sample} where the count of items equals the answer.
We further divide questions in \finqa~into four kinds: \textit{Span}, \textit{Spans}, \textit{Arithmetic} and \textit{Counting}, where the latter two kinds correspond to the above two types of deviations, to help us better investigate the numerical reasoning capability of a QA model. 

\noindent \textbf{Answer Source Annotation}
For each answer, annotators are required to specify the source(s) it is derived from, including \textit{Table}, \textit{Text}, and \textit{Table-text} (both).
This is to force the model to learn to aggregate information from hybrid sources to infer the answer, thus lift its generalizability.
For example, to answer Question 7 in \figref{sample}: \emph{``How much does the commercial cloud revenue account for the total revenue in 2019?''}, we can observe from the derivation that ``$125,843$ million'' comes from the table while ``$38.1$ billion'' from text.

\subsection{Quality Control}
To ensure the quality of annotation in \finqa, we apply strict quality control procedures.

\noindent \textbf{Competent Annotators} 
To build \finqa, financial domain knowledge is necessary.
Hence, we employ about $30$ university students majored in finance or similar disciplines as annotators.
We give all candidate annotators a minor test and only those with $95\%$ correct rate are hired.
Before starting the annotation work, we give a training session to the annotators to help them fully understand our annotation requirements and also learn the usage of our annotation system.

\noindent \textbf{Two-round Validation}
For each annotation, we ask two different verifiers to perform a two-round validation after it is submitted, including checking and approval, to ensure its quality.
We have five verifiers in total, including two annotators who have good performance on this project and three graduate students with financial background.
In the checking phase, a verifier checks the submitted annotation and asks the annotator to fix it if any mistake or problem is found.
In the approval phase, a different verifier inspects the annotation again that has been confirmed by the first verifier, and then approves it if no problem is found.

\subsection{Dataset Analysis}
Averagely, an annotator can label two hybrid contexts per hour; the whole annotation work lasts about three months.
Finally, we attain a total of $2,757$  hybrid contexts and $16,552$ corresponding question-answer pairs from $182$ financial reports.
The hybrid contexts are randomly split into training set ($80\%$), development set ($10\%$) and test set ($10\%$); hence all questions about a particular hybrid context belong to only one of the splits.
We show the basic statistics of each split in \tabref{finqa_stats}, and the question distribution regarding answer source and answer type in \tabref{type-source}.
In \figref{sample}, we give an example from \finqa, demonstrating the various reasoning types and percentage of each reasoning type over the whole dataset.

\begin{table}[htb]
\centering
\small
\begin{tabular}{@{}lrrrr@{}}
\toprule
{\bf Statistic} & {\bf Train} & {\bf Dev} & {\bf Test}  \\ 
\midrule
   \# of hybrid contexts & 2,201 & 278 & 278 \\
    \# of questions & 13,215 & 1,668 & 1,669 \\
    Avg. rows / table & 9.4 & 9.7 & 9.3 \\
    Avg. cols / table & 4.0 & 3.9 & 4.0 \\
    Avg. paragraphs / table & 4.8 & 4.9 & 4.6 \\
    Avg. paragraph len [words] & 43.6 & 44.8 & 42.6 \\
    Avg. question len [words] & 12.5 & 12.4 & 12.4 \\
    Avg. answer len [words] & 4.1 & 4.1 & 4.3 \\
    \bottomrule
    \end{tabular}
\caption{Basic statistics of each split in \finqa}
    \label{tab:finqa_stats}    
\end{table}

\begin{table}[t]
\centering
\small
\begin{tabular}{L{1.5cm}R{0.8cm}R{0.8cm}R{1.5cm}R{1cm}}
\toprule
& \bf Table & \bf Text & \bf Table-text & \bf Total \\ 
\midrule
Span & 1,801 & 3,496 & 1,842 & 7,139 \\
Spans & 777 & 258 & 1,037 & 2,072 \\
Counting & 106 & 5 & 266 & 377 \\
Arithmetic & 4,747 & 143 & 2,074 & 6,964 \\
Total & 7,431 & 3,902 & 5,219 & 16,552 \\
\bottomrule
\end{tabular}
\caption{Question distribution regarding different answer types and  sources in \finqa}
    \label{tab:type-source}    
\end{table}

%% file: 03_model.tex
\begin{figure*}[htb]
    \begin{center}
    \includegraphics[scale=0.65]{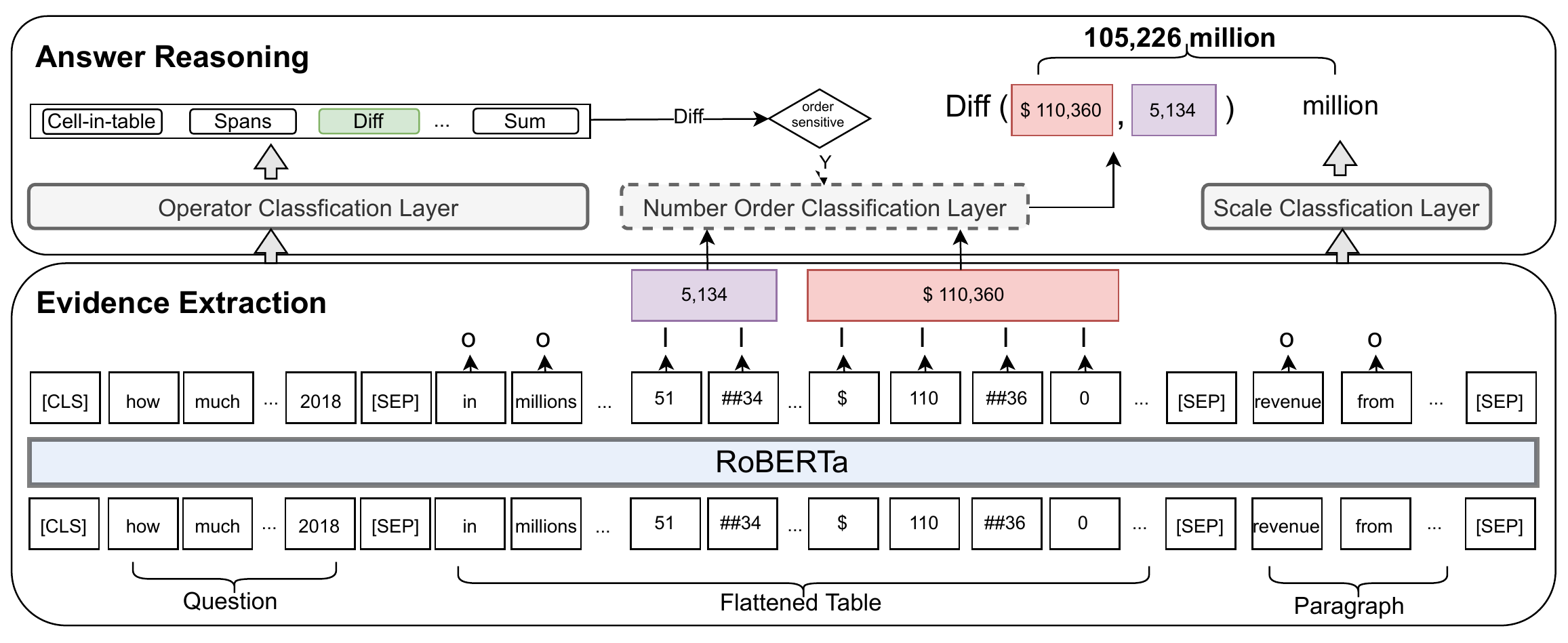}
    \end{center}
    \vspace{-1em}
    \caption{\label{fig:tagop} 
    Illustration of the architecture of proposed \tagop~model.
    Given 
     Question 6 in \figref{sample} where the hybrid context is also shown, 
    \tagop~supports $10$ \textit{operators}, which are described in Section~\ref{ops}.}
\end{figure*}

We introduce a novel QA model, named \tagop, which first applies sequence \underline{TAG}ging to extract relevant cells from the table and text spans from the paragraphs inspired by~\citep{li2016dataset,sun2016table,segal2020simple}. This step is analogy to slot filling or schema linking, whose effectiveness has been demonstrated in dialogue systems~\cite{lei2018sequicity, jin2018explicit} and semantic parsing~\cite{lei2020re}. And then \tagop~ performs symbolic reasoning over them with a set of aggregation \underline{OP}erators to arrive at the final answer.
The overall architecture is illustrated in \figref{tagop}.

\subsection{Sequence Tagging}
Given a question, \tagop~first extracts supporting evidences from its hybrid context (i.e. the table and associated paragraphs) via sequence tagging with the \underline{I}nside–\underline{O}utside tagging (\texttt{IO}) approach~\citep{ramshaw1995text}.
In particular, it assigns each token either \texttt{I} or \texttt{O} label and takes those tagged with \texttt{I} as the supporting evidences for producing the answer.
The given question, flattened table by row~\citep{herzig2020tapas} and associated paragraphs are input sequentially to a transformer-based encoder like RoBERTa~\citep{Liu2019RoBERTa}, as shown in the bottom part of \figref{tagop}, to obtain corresponding representations.
Each sub-token is tagged independently, and the corresponding cell in the table or word in the paragraph would be regarded as positive if any of its sub-tokens is tagged with \texttt{I}.
For the paragraphs, the continuous words that are predicted as positive are combined as a span.
During testing, all positive cells and spans are taken as the supporting evidences. 
Formally, for each sub-token $t$ in the paragraph, the probability of the tag is computed as
\begin{align}
    \mathbf{p}^\textrm{tag}_\textrm{t} &= \textrm{softmax}(\textrm{FFN}({h_t}))
\end{align}
where FFN is a two-layer feed-forward network with GELU~\citep{Dan2016GELU} activation and $h_t$ is the representation of sub-token~$t$.

\subsection{Aggregation Operator}
\label{ops}
Next, we perform symbolic reasoning over obtained evidences to infer the final answer, for which we apply an aggregation operator.
In our \tagop, there are ten types of aggregation operators. 
For each input question, an operator classifier is applied to decide which operator the evidences would go through; for some operators sensitive to the order of input numbers, 
an auxiliary number order classifier is used. 
The aggregation operators are explained as below, covering most reasoning types as listed in \figref{sample}.
\begin{itemize}[itemsep=1mm, parsep=0pt, leftmargin=*]
\item \textit{Span-in-text}: To select the span with the highest probability from predicted candidate spans. The probability of a span is the highest probability of all its sub-tokens tagged \texttt{I}.
\item \textit{Cell-in-table}: To select the cell with the highest probability from predicted candidate cells. The probability of a cell is the highest probability of all its sub-tokens tagged \texttt{I}.
\item \textit{Spans}: To select all the predicted cell and span candidates;
\item \textit{Sum}: To sum all predicted cells and spans purely consisting of numbers;
\item \textit{Count}: To count all predicted cells and spans;
\item \textit{Average}: To average over all the predicted cells and spans purely consisting of numbers;
\item \textit{Multiplication}: To multiply all predicted cells and spans purely consisting of numbers;
\item \textit{Division}: To first rank all the predicted cells and spans purely consisting of numbers based on their probabilities, and then apply division calculation to top-two;
\item \textit{Difference}: To first rank all predicted numerical cells and spans based on their probabilities, and then apply subtraction calculation to top-two.
\item \textit{Change ratio}: For the top-two values after ranking all predicted numerical cells and spans based on their probabilities, compute the change ratio of the first value compared to the second one. 
\end{itemize}

\noindent \textbf{Operator Classifier}
To predict the right aggregation operator, a multi-class classifier is developed.
In particular, we take the vector of \texttt{[CLS]} as input to compute the probability:
\begin{align}
    \mathbf{p}^\textrm{op} &= \textrm{softmax}(\textrm{FFN}(\texttt{[CLS]})
\end{align}
where FFN denotes a two-layer feed-forward network with the GELU activation. 

\noindent \textbf{Number Order Classifier} For operators of \textit{Difference}, \textit{Division} and \textit{Change ratio}, the order of the input two numbers matters in the final result.
Hence we additionally append a number order classifier after them, formulated as
\begin{align}
    \mathbf{p}^\textrm{order} &= \textrm{softmax}(\textrm{FFN}(\textrm{avg}(\texttt{$h_{t1}$},\texttt{$h_{t2}$}))
\end{align}
where FFN denotes a two-layer feed-forward network with the GELU activation, $h_{t1}$, $h_{t2}$ are representations of the top two tokens  according to probability, and
``avg'' means average.
For a token, its probability is the highest probability of all its sub-tokens tagged \texttt{I}, and its representation is the average over those of its sub-tokens.

\subsection{Scale Prediction}
Till now we have attained the string or numerical value to be contained in the final answer.
However, a right prediction of a numerical answer should not only include the right number but also the correct scale.
This is a unique challenge over \finqa~and very pervasive in the context of finance.
We develop a multi-class classifier to predict the scale.
Generally, the scale in \finqa~may be \textit{None}, \textit{Thousand}, \textit{Million}, \textit{Billion}, and \textit{Percent}.
Taking as input the concatenated representation of \texttt{[CLS]}, the table and paragraphs sequentially, the multi-class classifier computes the probability of the scale~as 
\begin{align}
    \mathbf{p}^\textrm{scale} &= \textrm{softmax}(\textrm{FFN}([\texttt{[CLS]};h_{tab};h_p])
    \label{eq_scale}
\end{align}
where $h_{tab}$ and $h_p$ are the representations of the table and the paragraphs respectively, which are obtained by applying an average pooling over the representations of their corresponding tokens,``;'' denotes concatenation, and FFN denotes a two-layer feed-forward network with the GELU activation. 

After obtaining the scale, the numerical or string prediction is multiplied or concatenated with the corresponding scale as the final prediction to compare with the ground-truth answer respectively.

\subsection{Training}

To optimize \tagop, the overall loss is the sum of the loss of the above four classification tasks:
\begin{equation}
\begin{split}
    \textrm{$\mathcal{L}$} =~ & \textrm{NLL}(\mathbf{log(P}^\textrm{tag}), \mathbf{G}^\textrm{tag})~\textrm{+} \\
    &\textrm{NLL}(\mathbf{log(P}^\textrm{op}),  \mathbf{G}^\textrm{op})~\textrm{+} \\
    &\textrm{NLL}(\mathbf{log(P}^\textrm{scale}), \mathbf{G}^\textrm{scale})~\textrm{+}  \\
    & \textrm{NLL}(\mathbf{log(P}^\textrm{order}), \mathbf{G}^\textrm{order})\\
    \end{split}
\end{equation}
where NLL(·) is the negative log-likelihood loss, $\mathbf{G}^\textrm{tag}$ and $\mathbf{G}^\textrm{op}$ come from the supporting evidences which are extracted from the annotated answer and derivation.
We locate the evidence in the table first if it is among the answer sources, and otherwise in its associated paragraphs. 
Note we only keep the first found if an evidence appears multiple times in the hybrid context.
$\mathbf{G}^\textrm{scale}$ uses the annotated scale of the answer;
$\mathbf{G}^\textrm{order}$ is needed when the ground-truth operator is one of \textit{Difference}, \textit{Division} and \textit{Change ratio}, which is obtained by mapping the two operands extracted from their corresponding ground-truth deviation in the input sequence.
If their order is the same as that in the input sequence, $\mathbf{G}^\textrm{order} = 0$; otherwise it is $1$.

%% file: 04_experiment.tex
\subsection{Baselines}
\label{sec:baselines}
\noindent \textbf{Textual QA Models}
We adopt two reading comprehension (RC) models as baselines over textual data: BERT-RC~\citep{Devlin2018Bert}, which is a SQuAD-style RC model;
and NumNet+ V2~\footnote{https://github.com/llamazing/numnet\_plus}~\citep{ran2019numnet}, which achieves promising performance on DROP that requires numerical reasoning over textual data.
We adapt them to our \finqa~as follows.
We convert the table to a sequence by row, also as input to the models, followed by tokens from the paragraphs. 
Besides, we add a multi-class classifier, exactly as in our \tagop, to enable the two models to predict the scale based on Eq.~\eqref{eq_scale}.

\noindent \textbf{Tabular QA Model}
We employ \tapas~for WikiTableQuestion (WTQ)~\citep{herzig2020tapas} as a baseline over tabular data.
\tapas~is pretrained over large-scale tables and associated text from Wikipedia jointly for table parsing.
To train it, we heuristically locate the evidence in the table with the annotated answer or derivation, which is the first matched one if a same value appears multiple times.
In addition, we remove the ``numerical rank id'' feature in its embedding layer, which ranks all values per numerical column in the table but does not make sense in \finqa. 
Similar to above textual QA setting, we add an additional multi-class classifier to predict the scale as in Eq.~\eqref{eq_scale}.

\noindent \textbf{Hybrid QA Model}
We adopt HyBrider~\citep{chen2020hybridqa} as our baseline over hybrid data, which tackles tabular and textual data from Wikipedia.
We use the code released in the original paper\footnote{https://github.com/wenhuchen/HybridQA}, but adapt it to \finqa. 
Concretely, each cell in the table of \finqa~is regarded as ``linked'' with associated paragraphs of this table, like hyperlinks in the original paper, and we only use its cell matching mechanism to link the question with the table cells in its linking stage.
The selected cells and paragraphs are fed into the RC model in the last stage to infer the answer.
For ease of training on \finqa, 
we also omit the prediction of the scale, i.e. 
we regard the predicted scale by this model as always correct.

\subsection{Evaluation Metrics}
We adopt the popular Exact Match (EM) and numeracy-focused \fone{} score \citep{Dua2019DROP} to measure model performance on \finqa. 
However, the original implementation of both metrics is insensitive to whether a value is positive or negative in the answer as the minus is omitted in evaluation.  
Since this issue is crucial for correctly interpreting numerical values, especially in the finance domain, we keep the plus-minus of a value when calculating them.
In addition, the numeracy-focused \fone{} score is set to $0$ unless the predicted number multiplied by predicted scale equals exactly the ground truth.

\subsection{Results and Analysis}

In the following, we report our experimental results on dev and test sets of \finqa.

\noindent \textbf{Comparison with Baselines}
We first compare our \tagop~with three types of previous QA models as described in Section~\ref{sec:baselines}.
The results are summarized in \tabref{overall-metric}.
It can be seen that our model is always superior to other baselines in terms of both metrics, with very large margins over the second best, namely  50.1/58.0 vs. 37.0/46.9 in EM/\fone{} on test set of \finqa~respectively.
This well reveals the effectiveness of our method that reasons over both tabular and textual data involving lots of numerical contents. 
For two textual QA baselines, NumNet+ V2 performs better than BERT-RC, which is possibly attributed to the stronger capability of numerical reasoning of the latter, but it is still worse than our method.
The tabular QA baseline Tapas for WTQ is trained with only tabular data in \finqa, showing very limited capability to process hybrid data, as can be seen from its performance.
The HyBrider is the worst among all baseline models, because it is designed for HybridQA~\citep{chen2020hybridqa} which does not focus on the comprehensive interdependence of table and paragraphs, nor  numerical reasoning.

However, all the models perform significantly worse than human performance\footnote{The human performance is evaluated by asking annotators to answer 50 randomly sampled hybrid contexts (containing 301 questions) from our test set. 
Note the human performance is still not 100\% correct because our questions require relatively heavy cognitive load like tedious numerical calculations. Comparing human performance of \fone{} in SQUAD~\citep{Rajpurkar2016SQuAD} ($86.8$\%) and DROP~\citep{Dua2019DROP}) ($96.4$\%), the score ($90.8$\%) in our dataset already indicates a good quality and annotation consistency in our dataset.}, indicating \finqa~is challenging to current QA models and more efforts on hybrid QA are demanded.

\noindent \textbf{Answer Type and Source Analysis}
Furthermore, we analyze detailed performance of \tagop~w.r.t answer type and source ~in \tabref{detail-metric}.
It can be seen that \tagop~performs better on the questions whose answers rely on the tables compared to those from the text.
This is probably because table cells have clearer boundaries than text spans to the model, thus it is relatively easy for the model to extract supporting evidences from the tables leveraging sequence tagging techniques.
In addition, \tagop~performs relatively worse on arithmetic questions compared with other types.
This may be because the calculations for arithmetic questions are diverse and harder than other types, indicating the challenge of \finqa, especially for the requirement of numerical reasoning.

\begin{table}
    \small
    \centering
    \begin{tabular}{lrrrr}
    \toprule
     \multirow{2}{*}{\bf Method}    & \multicolumn{2}{c}{\bf Dev} & \multicolumn{2}{c}{\bf Test} \\
     \cmidrule(lr){2-3}
     \cmidrule(lr){4-5}
         & EM & \fone{} & EM&  \fone{}\\
    \midrule
    
    \bf Human  &   -   &   -   &    84.1   &   90.8   \\
    \addlinespace
        
    \multicolumn{5}{l}{\bf Textual QA}\\
    BERT-RC   &      9.5 & 17.9 & 9.1 & 18.7  \\
    NumNet+ V2   &  38.1 & 48.3 & 37.0 & 46.9 \\
    \addlinespace
    
    \multicolumn{5}{l}{\bf Tabular QA}\\
    \tapas~for WTQ    &  18.9 & 26.5 & 16.6 & 22.8 \\
    \addlinespace

    \multicolumn{5}{l}{\bf Hybrid QA}\\
     HyBrider    &  6.6 & 8.3 & 6.3 & 7.5 \\

    \addlinespace

     \bf \tagop & \bf 55.2 & \bf 62.7 & \bf 50.1 & \bf 58.0 \\
    \bottomrule
    \end{tabular}
    \caption{Performance of different models on dev and test set of \finqa. Best results are marked in bold.}
   
    \label{tab:overall-metric}
\end{table}

\begin{table}[h]
\centering
\footnotesize
\begin{tabular}{L{1.2cm}C{1.0cm}C{1.0cm}C{1.3cm}C{1.0cm}}
\toprule
& Table & Text & Table-text & Total \\
\cmidrule{2-5}
& EM/\fone{} & EM/\fone{} & EM/\fone{} & EM/\fone{}
\\
\midrule
Span &  56.5/57.8 & 45.2/70.6 & 68.2/71.7 & 54.1/67.9 \\
Spans & 66.3/77.0 & 19.0/59.1 & 63.2/76.9 & 60.0/75.1 \\
Counting & 63.6/63.6 & -/- & 62.1/62.1 & 62.5/62.5 \\
Arithmetic & 41.1/41.1 & 27.3/27.3 & 46.5/46.5 & 42.5/42.5 \\
Total &  47.8/49.3 & 43.3/68.7 & 58.3/62.2 & 50.1/58.0 \\
\bottomrule
\end{tabular}
\caption{Detailed experimental results of \tagop~w.r.t. answer types and sources on test set.
}

    \label{tab:detail-metric}    
\end{table}

\noindent \textbf{Results of \tagop~with Different Operators}
We here investigate the contributions of the ten aggregation operators to the final performance of \tagop.
As shown in \tabref{op-ablation}, we devise nine variants of the full model of \tagop; based on the variant of \tagop~with only one operator (e.g. Span-in-text), for each of other variants, we add one more operator back.
As can be seen from the table, all added operators can benefit the model performance.
Furthermore, we find that some operators like \emph{Span-in-text}, \emph{Cell-in-table}, \emph{Difference} and \emph{Average} make more contributions than others.
In comparison, \emph{Sum} and \emph{Multiplication} bring little gain or even decline.
After analysis, we find this is because the instances of \emph{Sum} or \emph{Multiplication} are minor in our test set, which are easily influenced by randomness.

\begin{table}
    \small
    \centering
    \begin{tabular}{lrrrr}
    \toprule
     \multirow{2}{*}{\bf Model}    & \multicolumn{2}{c}{\bf Dev} & \multicolumn{2}{c}{\bf Test} \\
     \cmidrule(lr){2-3}
     \cmidrule(lr){4-5}
         & EM & \fone{} & EM&  \fone{}\\
    \midrule
    + Span-in-text & 13.4 & 20.5 & 14.1 & 21.8 \\
    + Cell-in-table & 25.4 & 36.0 & 24.1 & 35.3 \\
    + Spans & 33.6 & 41.3 & 31.3 & 39.4 \\
    + Sum & 33.8 & 41.3 & 31.2 & 39.1 \\
    + Count & 35.9 & 43.5 & 32.7 & 40.6 \\
    + Average & 43.3 & 50.6 & 38.2 & 45.9 \\
    + Multiplication & 44.2 & 51.4 & 37.9 & 46.0 \\
    + Division & 45.0 & 52.5 & 39.2 & 47.5 \\
    + Difference & 51.4 & 58.7 & 45.1 & 53.3 \\
    + Change ratio (Full) & \bf 55.2 & \bf 62.7 & \bf 50.1 & \bf 58.0 \\
    \bottomrule
    \end{tabular}
    \caption{Performance with different aggregation operators of \tagop~model.}
    \label{tab:op-ablation}
\end{table}

\noindent \textbf{Error Analysis}
We further investigate our \tagop~by analysing error cases. 
We randomly sample $100$ error instances from the test set, and classify them into five categories as shown in \tabref{error-case}, each with an example:
(1) \emph{Wrong Evidence} ($55$\%), meaning the model obtained wrong supporting evidence from the hybrid context;
(2) \emph{Missing Evidence} ($29$\%), meaning the model failed to extract the supporting evidence for the answer;
(3) \emph{Wrong Calculation} ($9$\%), meaning the model failed to compute the answer with the correct supporting evidence;
(4) \emph{Unsupported Calculation} ($4$\%), meaning the ten operators defined cannot support this calculation;
(5) \emph{Scale Error} ($3$\%), meaning the model failed to predict the scale of the numerical value in an answer.

We can then observe about $84$\% error is caused by the failure to extract the supporting evidence from the table and paragraphs given a question.
This demonstrates more efforts are needed to strengthen the model's capability of precisely aggregating information from hybrid contexts.

After instance-level analysis, we find another interesting error resource is the dependence on domain knowledge. 
While we encourage annotators to create questions answerable by humans without much finance knowledge, we still find domain knowledge is required for some questions.
For example, given the question \emph{``What is the gross profit margin of the company in 2015?''},  the model needs to extract the gross profit and revenue from the hybrid context and compute the answer according to the finance formula \emph{(``gross profit margin = gross profit / revenue'')}. 
How to integrate such finance knowledge into QA models to answer questions in \finqa~still needs further exploration.

\begin{table}[h]
\centering
\footnotesize
\begin{tabular}{L{1.6cm}|L{5.2cm}}
\toprule
\multirow{3}{*}{ \makecell[l]{Wrong\\ Evidence\\(55\%)}} & Q: How much did the level 2 OFA change by from 2018 year end to 2019 year end? \\
& G: \textcolor{blue}{375} - 2,032 \\
& P: \textcolor{red}{1,941} - 2,032 \\
\midrule
\multirow{3}{*}{ \makecell[l]{Missing\\ Evidence\\(29\%) }} &  Q: How many years did adjusted EBITDA exceed \$4,000 million? \\
& G: \emph{count}(2017, 2018, \textcolor{blue}{2019}) \\
& P: \emph{count}(2017, 2018) \\
\midrule
\multirow{3}{*}{ \makecell[l]{Wrong\\Calculation \\
(9\%)} }
& Q: What is the change in the \% of pre-tax loss from 2018 to 2019? \\
& G: \textcolor{blue}{39\% - 20\%} \\
& P: \textcolor{red}{20\% - 39\%} \\
\midrule
\multirow{3}{*}{ \makecell[l]{Unsupported\\Calculation \\
(4\%)} } & Q: What is the proportion of investor relations and consultants over the total operating expense in 2019? \\
& G: \textcolor{blue}{(105,639 + 245,386) /19,133,139} \\
& P: \textcolor{red}{245,386 / 19,133,139} \\
\midrule

\multirow{3}{*}{\makecell[l]{ Scale\\Error\\(3\%)} } & Q: What is the closing price in March, 2020? \\
& G: 0.22 \\
& P: 0.22 \textcolor{red}{million} \\
\bottomrule
\end{tabular} 
\caption{Examples of error and corresponding percentage. 
Q, G, P denote question, ground truth, prediction.}
\label{tab:error-case} 
\end{table}

%% file: 05_related_work.tex
\noindent \textbf{QA Datasets}
Currently, there are many datasets for QA tasks, focusing on text, or KB/table.
Textual ones include CNN/Daily Mail~\cite{Hermann2015Teaching}, SQuAD \cite{Rajpurkar2016SQuAD}, etc.
Recently deep reasoning over textual data has gained increasing attention~\cite{Zhu2021Retrieving}, e.g. multi-hop reasoning~\citep{yang2018hotpotqa,Welbl2018Constructing}. 
DROP~\citep{Dua2019DROP} is built to develop numerical reasoning capability of QA models, which in this sense is similar to \finqa, but only focuses on textual data.
KB/Tabular QA aims to automatically answer questions via well-structured KB~\citep{berant2013semantic,talmor2018web,yih2015semantic} or semi-structured tables~\citep{Pasupat2015Compositional,zhongSeq2SQL2017,yu2018spider}.
Comparably, QA over hybrid data receives limited efforts, focusing on mixture of KB/tables and text.
HybridQA~\citep{chen2020hybridqa} is one existing hybrid dataset for QA tasks, where the context is a table connected with Wiki pages via hyperlinks. 

\noindent \textbf{Numerical Reasoning} 
Numerical reasoning is key to many NLP tasks like question answering~\citep{Dua2019DROP,ran2019numnet, andor2019giving,chen2020question,Pasupat2015Compositional,herzig2020tapas, yin2020tabert,Zhang2020WTE} and arithmetic word problems~\citep{kushman2014learning,mitra2016learning,huang2017learning,ling2017program}.
To our best knowledge, no prior work attempts to develop models able to perform numerical reasoning over hybrid contexts.

%% file: 06_conclusion.tex
We propose a new challenging QA dataset \finqa, comprising real-word hybrid contexts where the table contains numbers and has comprehensive dependencies on text in finance domain. To answer questions in \finqa, the close relation between table and paragraphs and numerical reasoning are required.
We also propose a baseline model \tagop~based on \finqa, aggregating information from hybrid context and performing numerical reasoning over it with pre-defined operators to compute the final answer.
Experiments show \finqa~dataset is very challenging and more effort is demanded for tackling QA tasks over hybrid data.
We expect our \finqa~dataset and \tagop~model would serve as a benchmark and baseline respectively to help build more advanced QA models, facilitating the development of QA technologies 
to address more complex and realistic hybrid data, especially those requiring numerical reasoning.

%% file: 07_appendix.tex
\subsection{Table Analysis}
To maintain the semi-structured nature of financial tables, we almost keep the same table structure in \finqa~as that in the original financial reports.
We sample $100$ hybrid contexts from the training set to conduct a manual evaluation to assess the complexity of the table structures.
Specifically, we analyze the distribution w.r.t. the number of \textit{row headers}, as shown in \tabref{header}.
It can be seen that around $79\%$ of the tables have two or more row-headers, indicating large difficulty in interpreting financial tables.
In addition, we have also found that all sampled tables all have one \textit{column header}.

\begin{table}[h]
\centering
\begin{tabular}{c|r}
\toprule
\bf \# of Row Header &  \bf Proportion (\%) \\
\midrule
1 & 21 \\
2 & 68 \\
3 & 9 \\
more than 3 & 2 \\
\bottomrule
\end{tabular}
\caption{Distribution of no. of row-headers in \finqa.}
    \label{tab:header}    
\end{table}

\subsection{Operator Classifier}
We present the proportion of questions that should go through each aggregation operator (ground truth), as well as the performance of our operator classifier on dev and test set in \tabref{op}.

\begin{table}[htb]
\centering
\begin{tabular}{L{2cm}rrrr}
\toprule
\multirow{2}{*}{ \bf Operator } & \multicolumn{2}{c}{ \bf Dev } & \multicolumn{2}{c}{ \bf Test } \\
 \cmidrule(lr){2-3}
 \cmidrule(lr){4-5}
& \% & \bf Acc & \% & \bf Acc \\
\midrule
Span-in-text & 20.9 & 92.3 & 21.3 & 91.6 \\
Cell-in-table & 21.1 & 91.2 & 21.6 & 86.7 \\
Spans & 13.0 & 96.8 & 12.6 & 93.8 \\
Sum & 3.4 & 86.0 & 2.5 & 76.2 \\
Count & 1.9 & 93.8 & 2.4 & 100.0 \\
Average & 8.5 & 100.0 & 5.9 & 100.0 \\
Multiplication & 0.2 & 33.3 & 0.1 & 0.0 \\
Division & 1.0 & 76.5 & 1.0 & 87.5 \\
Difference & 14.1 & 96.6 & 15.9 & 96.6 \\
Change ratio & 9.3 & 96.1 & 10.2 & 95.3 \\
Other & 6.6 & 0.0 & 6.6 & 0.0 \\
\bottomrule
\end{tabular}

\caption{Ground truth proportion of questions that should be fed to different operators and prediction accuracy by operator classifier of \tagop~on dev and test set of \finqa.}
    \label{tab:op}    
\end{table}

\subsection{Scale Prediction}
We report the proportion of the ground truth scale in an answer and also the performance of our scale predictor on dev and test set in \tabref{scale-pp}.

\begin{table}[htb]
\centering
\begin{tabular}{L{2cm}rrrr}
\toprule
\multirow{2}{*}{ \bf Scale } & \multicolumn{2}{c}{ \bf Dev } & \multicolumn{2}{c}{ \bf Test } \\
 \cmidrule(lr){2-3}
     \cmidrule(lr){4-5}
& \% & \bf Acc & \% & \bf Acc \\
\midrule
None & 47.6 & 92.4 & 50.3 & 90.1 \\
Thousand & 20.7 & 96.8 & 19.2 & 95.3 \\
Million & 15.2 & 92.1 & 12.9 & 90.2 \\
Billion & 0.4 & 28.6 & - & - \\
Percent & 16.1 & 95.9 & 17.7 & 95.9 \\
\bottomrule
\end{tabular}

\caption{The proportion of ground truth scale on dev and test set of \finqa~with prediction accuracy by scale predictor of \tagop.}
    \label{tab:scale-pp}    
\end{table}